# The PAR dataset: Prostate biopsy whole slide images from an underrepresented Middle Eastern population

Peshawa J. Muhammad Ali[1,2*], Navin Vincent[3*], Saman S. Abdulla[4,5], Han N. Mohammed Fadhl[6], Anders Blilie[7,8], Kelvin Szolnoky[9], Julia Anna Mielcarz[3], Xiaoyi Ji[9], Kimmo Kartasalo[3,10,11], Abdulbasit K. Al-Talabani[1ǂ], Nita Mulliqi[3ǂ]

1. Department of Software Engineering, Faculty of Engineering, Koya University, Koya 44023, Kurdistan Region, Iraq
2. Department of Mechanical and Manufacturing Engineering, Faculty of Engineering, Koya University, Koya 44023, Kurdistan Region, Iraq
3. Department of Medical Epidemiology and Biostatistics, SciLifeLab, Karolinska Institutet, Stockholm, Sweden
4. College of Dentistry, Hawler Medical University, Erbil, Kurdistan Region, Iraq
5. PAR Private Hospital, Erbil, Kurdistan Region, Iraq
6. College of Dentistry, University of Sulaimani, Sulaymaniyah, Kurdistan Region, Iraq
7. Department of Pathology, Stavanger University Hospital, Stavanger, Norway
8. Faculty of Health Sciences, University of Stavanger, Stavanger, Norway
9. Department of Medical Epidemiology and Biostatistics, Karolinska Institutet, Stockholm, Sweden
10. ELLIS Institute Finland, Espoo, Finland
11. Department of Computing, Faculty of Technology, University of Turku, Turku, Finland

* These authors contributed equally to this work.
ǂ These authors jointly supervised the work.
Corresponding author: nita.mulliqi@ki.se.

## Abstract

Artificial intelligence (AI) is increasingly used in digital pathology. Publicly available histopathology datasets remain scarce, and those that do exist predominantly represent Western populations. Consequently, the generalizability of AI models to populations from less digitized regions, such as the Middle East, is largely unknown. This motivates the public release of our dataset to support the development and validation of pathology AI models across globally diverse populations. We present 1,017 whole slide images by digitizing 339 glass slides of prostate core needle biopsies from a consecutive series of 185 patients collected in Erbil, Iraq. Each glass slide was scanned by three different whole-slide scanners. The dataset also includes the corresponding Gleason Scores and the International Society of Urological Pathology grades assigned independently by three pathologists. All slides were de-identified and are provided in their native formats without further conversion. The dataset enables grading concordance analyses, color normalization, and cross-scanner robustness evaluations. The dataset is publicly available through the BioImage Archive and released under the CC BY 4.0 license.

## Background & summary

Prostate cancer diagnosis relies on microscopic assessment of hematoxylin and eosin (H&E) stained biopsy slides. In routine practice, core needle biopsy is the most common method for diagnosing prostate cancer, where a needle is used to take a small tissue sample from the prostate, guided by ultrasound and potentially MRI. After extraction, tissue cores are processed through routine histopathological steps to produce a glass slide for each biopsy. Each slide is subsequently reviewed and graded by a pathologist according to the Gleason scoring (GS) system [1]. The GS consists of primary and secondary grades based on the proportion of tissue with different Gleason patterns, e.g., a score of 4 + 3 = 7, where the first number represents the predominant pattern [2]. In 2014, the International Society of Urological Pathology (ISUP) introduced the ISUP grades, pooling GS into five ordinal categories, from 1 to 5, with grade 1 being the least aggressive and grade 5 the most aggressive cancer [3]. Both GS and ISUP grades remain in parallel use in clinical practice. Despite the central clinical role of Gleason grading, its reproducibility remains limited. Numerous studies have demonstrated substantial intra- and inter-pathologist variability leading to under- and over-treatment of patients [4–6]. Digital pathology and artificial intelligence (AI)-based analysis of whole-slide images (WSIs) offer opportunities to improve diagnostic consistency [7], and many studies now show that AI systems can diagnose and grade prostate cancer with performance comparable to expert pathologists [8–11].

To further accelerate AI development, several efforts have been made to release publicly available digitized prostate pathology datasets. The PANDA dataset released 5,160 digitized core needle biopsies from the Netherlands and 5,456 from Sweden with associated ISUP grades and, in some cases, tissue masks annotating the cancerous regions [12]. The SPROB20 dataset released 2,611 digitized biopsies collected in Sweden with associated ISUP and other clinical variables [13]. The SICAPv2 published 155 digitized biopsies collected in Spain with associated GS, provided as .jpg patches instead of the full WSIs [14,15]. The UKK and WNS datasets comprise 100 digitized biopsies collected in Germany and Austria, with associated ISUP grades from 10 different pathologists. These two datasets are ISUP-balanced and do not contain benign samples [16]. The DiagSet dataset includes digitized prostate biopsies collected in Poland as (i) 2.6 million patches with patch-level custom labels (e.g., benign, cancer of lower/higher severity), (ii) 4,675 WSIs with cancer/non-cancer labels, and (iii) 46 WSIs labeled with cancer/non-cancer/ambiguous (requires immunohistochemistry (IHC)) labels provided by nine pathologists [17]. The Gleason grading challenge 2022 includes 144 digitized prostatectomies and 53 biopsies collected in Singapore, with associated Gleason pattern pixel-level annotations [18]. The Cancer Genome Atlas Prostate Adenocarcinoma (TCGA-PRAD) project released 499 digitized prostatectomies collected in the US, accompanied by GS, genomics, and other clinical data [19,20]. The EMPaCT tissue microarray (TMA) dataset comprises 210 digitized TMAs collected in Switzerland, associated with binary labels for overall survival and disease progression status. It also includes consecutive tissue sections of H&E staining and additional immunohistochemical markers. The authors released the tissue cores as .png images [21]. Zhong et al. [22] introduced a Swiss dataset derived from 73 unique patients comprising 71 TMA cores, two WSIs and 6,000 derived patches, with consensus annotations from two pathologists. The Tissue Microarrays Zürich (TMAZ) includes five prostate cancer TMAs, each containing approximately 200–300 cores from prostatectomy specimens collected in Switzerland, each associated with Gleason pattern annotations from two pathologists [23]. The Gleason 2019 Challenge dataset released 331 TMA cores collected in Canada as .jpg images [24] with six separate expert annotation maps. In addition to these publicly available resources, many AI studies rely on proprietary industry datasets [8] or large single-institution archives [25, 26], which are not publicly accessible.

Despite efforts in the field to publish publicly available data, challenges still remain. First, given the substantial variability in Gleason grading, studies need to rely on datasets that provide reference grading from more than one pathologist. The available datasets typically contain single reference standards. This severely limits reproducibility and benchmarking. Secondly, most AI studies are developed and validated predominantly in Caucasian populations. This raises concerns about model generalization to underrepresented regions where pathology services are not yet digitized. Full workflow digitization requires substantial financial investment, limiting its feasibility in low-resource or low-volume settings. This represents a chicken-and-egg problem—lack of digital pathology precludes the collection of data and evidence for the effectiveness and generalizability of AI pathology in these populations, and the absence of this evidence in part contributes to the lack of investment in digitization. One potential solution would be to use cost-effective compact scanners for validation and pilot testing of AI in currently non-digital settings. No public datasets currently include data digitized with compact scanners or originating from underrepresented populations. Thirdly, numerous studies have demonstrated that AI model performance varies across whole-slide scanners and that scanner-induced differences impact model generalization [27,28]. Existing public datasets typically provide WSIs from only one scanner type, limiting the ability to study cross-scanner robustness.

Together, these gaps underscore the need for datasets that enable cross-population, multi-scanner, and multi-pathologist AI validation to support broad clinical adoption. To address these gaps, we present 1,017 WSIs of prostate core needle biopsies derived from 339 glass slides from 185 patients collected at the PAR Hospital in Erbil, Iraq. Each glass slide was digitized using three different scanners, resulting in three scan versions per glass slide. The consecutive patient series is representative of the cases encountered in routine clinical practice between 2013-2024. The slides are associated with slide-level GS and ISUP grades assigned independently by three pathologists blinded to each other and to any AI grading. The slides were digitized using three different whole-slide scanners: Leica Aperio GT 450 DX, Hamamatsu NanoZoomer HT 2.0, and the Grundium Ocus40 compact scanner. This work also addresses the broader benchmarking crisis in biomedical AI, where progress is constrained by limited public datasets, institution-specific biases, and inconsistent validation standards [29]. By making this dataset available, we aim to help bridge long-standing inequities in both population representation and technological access, moving the field closer to inclusive and reliable AI for pathology.

# Methods

## Ethical considerations

The patient sample collection was approved by the Ethical Committee of PAR Hospital in Kurdistan, Iraq (permit 1002/07072024). Digital scanning and processing of the samples at Karolinska Institutet was approved by the Swedish Ethical Review Authority (permit 2019-05220). The samples were de-identified before shipping to Karolinska Institutet. Informed consent was waived by the ethical review board due to the use of de-identified prostate specimens in a retrospective setting.

## Study design

This work consists of four main steps: slide selection, slide preparation in the laboratory, whole slide scanning, and annotation by pathologists to establish a reference standard.

**Slide selection**: The study cohort was derived from the pathology archive at the PAR Hospital (Erbil), which contained 30,056 cases assessed between 2013 and 2024. The archive reflects consecutive routine diagnostic cases. The study cohort was extracted by multiple manual steps.

An initial filtering of this digital archive, consisting of reports in .docx format, was performed using the keyword "pros". This primary search yielded 614 potentially relevant files. This subset underwent further cleaning to identify 502 relevant prostate cases, which included needle biopsies, transurethral resections of the prostate, and prostatectomies. For this study, the focus was narrowed to the 251 cases identified as "core needle biopsy files". This group was refined further by merging 12 pairs (24 files) identified as "addenda". These addenda files, which noted additional tests like IHC, were merged to avoid duplication, leaving 239 unique needle biopsy cases. The final eligibility criterion was the availability of the original formalin-fixed and paraffin-embedded (FFPE) tissue blocks in the lab archive. It was found that 54 tissue blocks were lost (mostly from the early years of the collection). After confirming eligibility, the final dataset consisted of 185 prostate cancer needle biopsy cases, from which new H&E slides were prepared from the existing blocks. In summary, the eligibility criteria were: (a) cases identified as prostate core needle biopsy, (b) non-duplicate/non-addenda reports, and (c) availability of the archival FFPE block (see **Figure 1**).

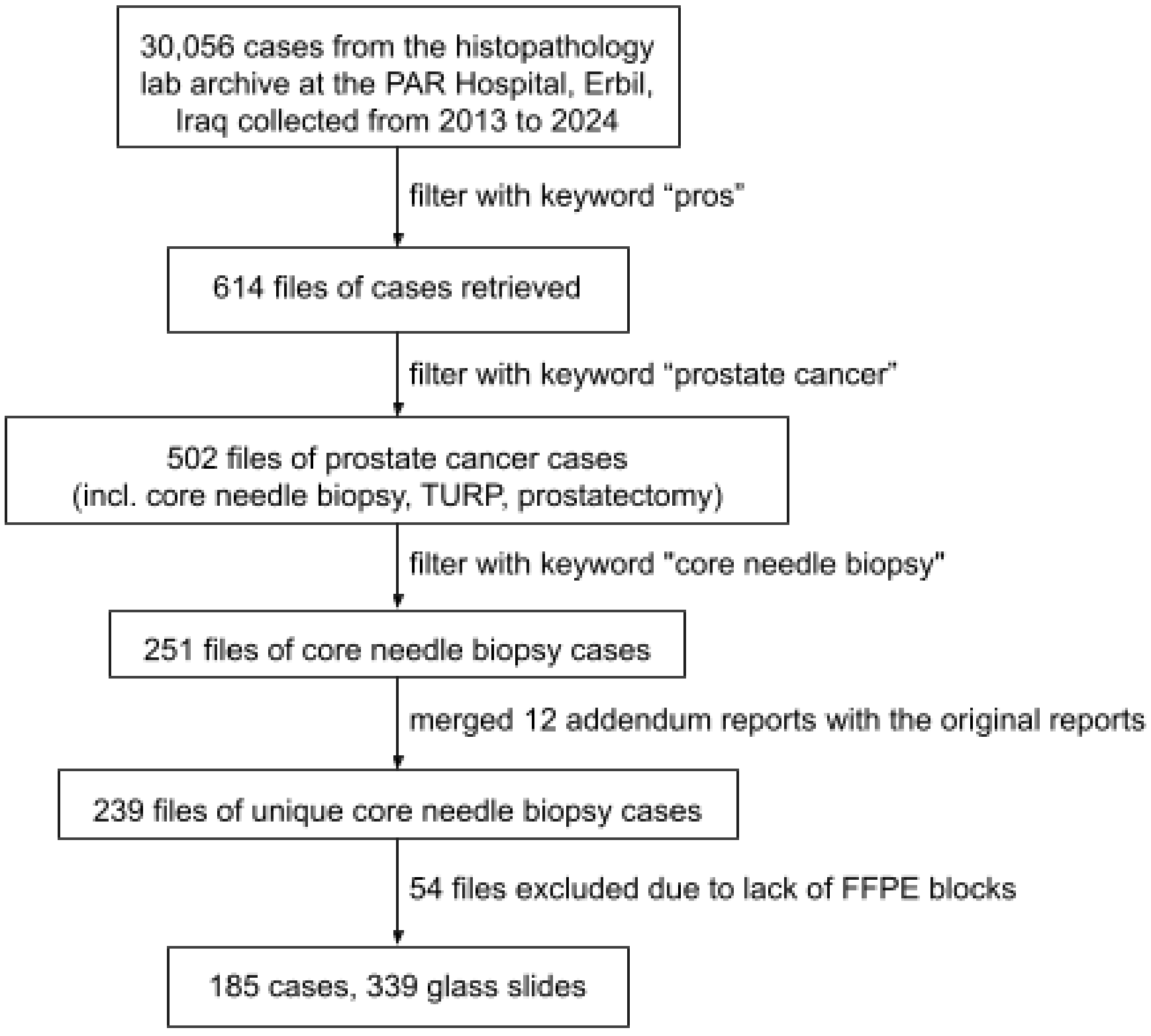

**Figure 1. Diagram illustrating dataset curation.** The diagram outlines the process of identifying 30,056 pathology cases from the PAR Hospital archive, filtering prostate-related

reports, removing addenda, confirming availability of archival FFPE blocks, and deriving the final set of 185 eligible prostate core needle biopsy cases for slide preparation and scanning.

**Slide preparation**: The standard biopsy protocol at the institution involved collecting six cores from each side of the prostate, though core fragmentation was occasionally observed during extraction. If biopsies from each side were submitted in separate containers, separate slides were generated. Conversely, if biopsies from both sides were submitted in a single container, they were combined to produce a single slide. After reception of the samples in the laboratory, they were placed in cassettes and underwent standard processing, including fixation in formalin to preserve morphology, dehydration through graded alcohols, clearing (often with xylene), and embedding in paraffin wax to form solid blocks (FFPE). These FFPE blocks were subsequently sectioned with a microtome into thin slices, typically 3–4 µm thick, with sections obtained at multiple levels to ensure adequate tissue representation. Sections were checked for quality, and additional deeper levels were obtained where it was deemed necessary by the pathologist. For 154 patients, biopsies from the left and right sides were submitted in separate containers, resulting in two slides per patient, whereas for the remaining 31 patients, both sides were submitted in a single container and therefore produced one slide per patient. The slides were stained with H&E and were finally coverslipped.

For this study, the original archived diagnostic H&E slides were not used, as many exhibited significant stain fading and degradation due to prolonged storage. Therefore, to ensure consistent and high-quality staining, all retrieved FFPE blocks were recut to achieve new sections, with prior re-embedding performed when deemed necessary, to generate fresh slides for analysis. The new slides were screened and assessed independently of the original pathology report.

**Whole slide image scanning**: All glass slides were scanned at the Department of Medical Epidemiology and Biostatistics, Karolinska Institutet, Stockholm, Sweden, using three whole-slide scanners, Hamamatsu NanoZoomer 2.0 HT, Leica Aperio GT 450 DX and a compact, single-slide scanner, Grundium Ocus40. Slides were digitized by trained personnel using an automated scanning workflow. Manual rescanning of slides was done in cases where automated focusing was not successful. All slides were scanned at 40×, with resolutions of 0.22 µm/pixel for Hamamatsu, 0.26 µm/pixel for Aperio, and 0.25 µm/pixel for Grundium. WSIs were stored in the scanners' proprietary formats: .ndpi for Hamamatsu and .svs for Grundium and Aperio. See **Figure 2** for example WSIs of the dataset [30].

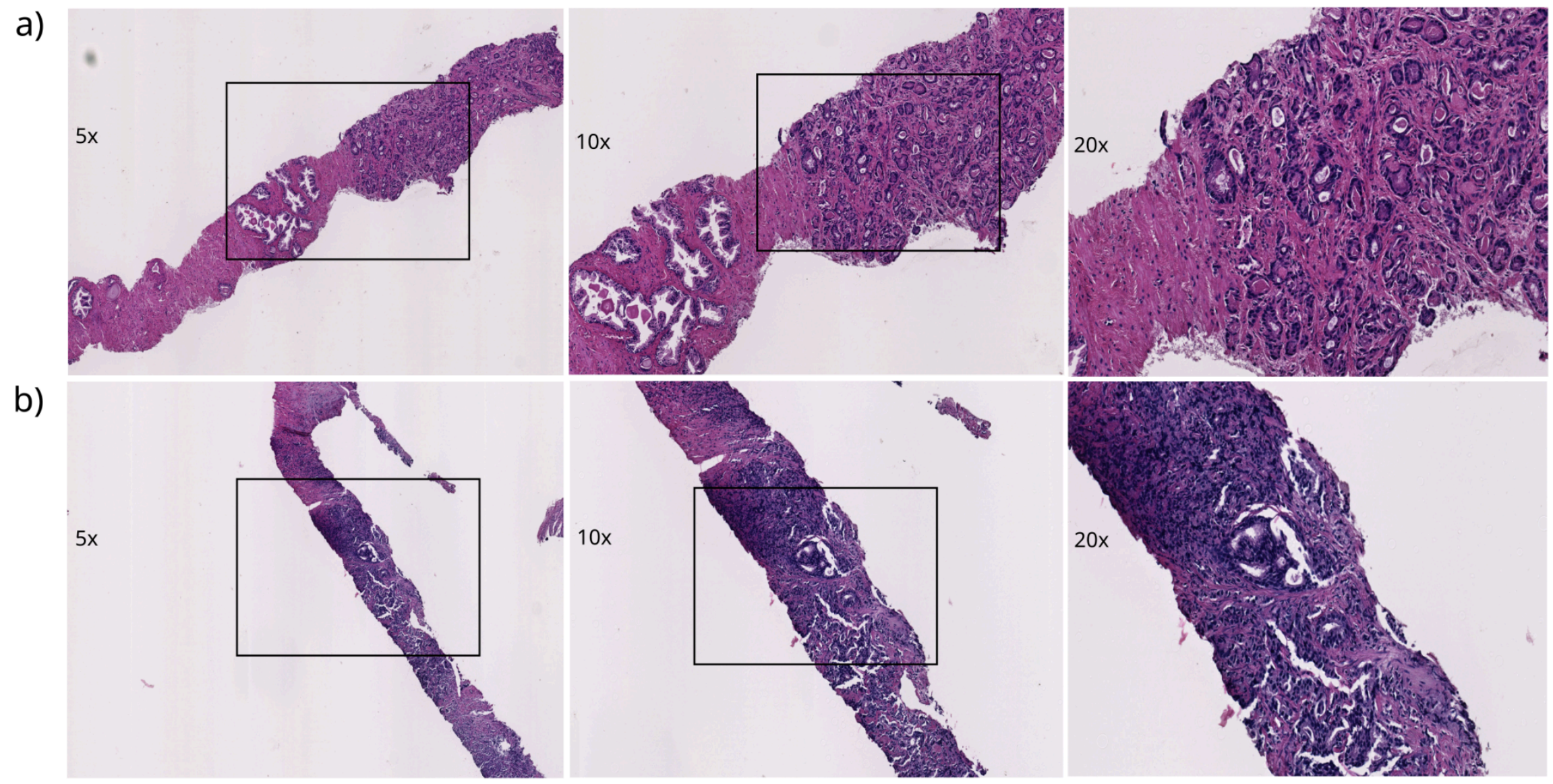

**Figure 2. Example whole slide images.** The figure illustrates two prostate core needle biopsy slides (a, b) digitized with the Hamamatsu NanoZoomer 2.0 HT whole slide scanner. For each slide, representative zoom-in regions highlight tissue morphology at higher magnification. The respective slides and regions were selected by the pathologist (H.M.). The maximum available magnification, 40x, corresponds to 0.2266 micrometers per pixel.

**Reference standard protocol**: The WSIs were independently assessed by three pathologists. The first was S.A., a general pathologist with 22 years of experience and a special interest in prostatic pathology. He works as a lecturer at Hawler Medical University and as a consultant pathologist at PAR Hospital in Erbil, Iraq, and serves as a senior examiner for the Royal College of Pathologists, UK. The second, H.M., is a general pathologist with 15 years of experience, currently working at the University of Sulaimani in Sulaymaniyah, Iraq. The third, A.B., is a uropathologist with 6 years of experience at Stavanger University Hospital in Stavanger, Norway. The pathologists assessed the slides digitally by using Cytomine [31] and assigned slide-level GSs (using 0+0 for benign slides). Scores were subsequently converted to ISUP grades and stored with the corresponding slide IDs. No pixel-level annotations for outlining regions of interest were marked during the assessment. The first pathologist (S.A.) graded all 339 slides, the second pathologist (H.M.) graded 337 slides, and the third uropathologist (A.B.) graded a random subset of 59 slides stratified by ISUP grades assigned by the first pathologist (see **Table 1**). Assessments were fully independent with no immediate discussion or reconciliation among pathologists. The distinction between uropathologists and general pathologists is not consistently defined across countries and clinics, and there are substantial differences in subspecialty training requirements and clinical responsibilities. This should be taken into account when interpreting agreement rates between general and specialized pathologists.

| **Number of patients** | | **Number of slides** | | | |
|---|---|---|---|---|---|
| **n = 185** | | **n = 339** | | | |
| **Age, years** | | **ISUP grade** | **Pathologist I** | **Pathologist II** | **Pathologist III** |
| **<= 49 yrs** | 5 | **Benign** | 164 | 208 | 11 |
| **50 – 54 yrs** | 9 | **ISUP 1 (3+3)** | 7 | 2 | 5 |
| **55 – 59 yrs** | 14 | **ISUP 2 (3+4)** | 38 | 7 | 13 |
| **60 – 64 yrs** | 26 | **ISUP 3 (4+3)** | 59 | 8 | 3 |
| **65 – 69 yrs** | 49 | **ISUP 4 (4+4, 3+5, 5+3)** | 30 | 25 | 6 |
| **>= 70 yrs** | 79 | **ISUP 5 (4+5, 5+4, 5+5)** | 41 | 86 | 21 |
| **Missing** | 3 | **Missing** | 0 | 3 | 280 |

**Table 1. Cohort characteristics.** The table outlines age distribution across the 185 patients, and ISUP grade distribution across 339 prostate biopsies from all pathologists.

## Data records

The PAR dataset [30] has been deposited to the BioImage Archive under accession number S-BIAD2323 (https://doi.org/10.6019/S-BIAD2323). The dataset is deposited as a single root folder, "iraq_prostate_wsi", containing six subfolders: "component_A_grundium_svs", "component_B_hamamatsu_ndpi", "component_C_leica_svs", annotations, "file_lists", and "docs". The three component folders each hold 339 WSIs collected from 185 patients, where 154 have two slides, and 31 have only one slide per patient. Filenames share a consistent, non-identifying scheme: *c*<slide_id><*a* | *b* | none>.<ext>, where *a* or *b* indicates when a patient has right/left slides, and is omitted when only one slide exists, and the extension reflects the scanner format. The annotations folder provides a single table that lists slide names and the associated Gleason scores and ISUP grades assigned independently by the three pathologists. GS were assigned using ten classes (0+0, 3+3, 3+4, 4+3, 4+4, 3+5, 5+3, 4+5, 5+4, 5+5), and ISUP grades show the GS grouped in 6 ordinal values as: ISUP 0 (GS 0+0), ISUP 1 (GS 3+3), ISUP 2 (GS 3+4), ISUP 3 (GS 4+3), ISUP 4 (GS 4+4 or 3+5 or 5+3), and ISUP 5 (GS 4+5, 5+4, or 5+5). The "file_lists" folder documents how files are linked or paired across scanners, and "docs" contains dataset documentation such as the license and README files. No personal identifiers are included; file and label names are sufficient for users to pair slides with labels and to link the three scanner versions of each slide.

## Technical validation

We verified technical quality through practical, user-facing checks:

(i) Completeness: each scanner component folder contains 339 WSIs, matching 185 patients, 154 with suffix *a* or *b*, and 31 single-slide cases.

(ii) Openability: tissue was segmented using deep-learning–based algorithms, and patches were generated from all WSIs. This process involved using OpenSlide [32] to access the WSIs and read the full image area. No issues were encountered when opening either file format (.svs or .ndpi). In addition, all WSIs were successfully viewed at multiple magnification levels in Cytomine during grading and separately in the Automated Slide Analysis Platform (ASAP) [33].

(iii) Annotation conformance: the annotations table uses a fixed dictionary with 10 GS classes and 6 ISUP grades.

## Usage notes

Traditional image viewers cannot read WSIs; native formats like .svs and .ndpi can be opened with specialized viewers and toolkits such as Cytomine [31], OpenSlide [32], ASAP [33], and QuPath [34]. While the annotation table does not contain a consensus label from the three pathologists, users can choose their own consensus rule explicitly, for instance, the majority, or the highest grade. The typical preprocessing for modeling includes foreground masking, tiling at different sizes, and stain/illumination normalization [35].

## Data availability

All data supporting this publication are openly available through the BioImage Archive (BIA) under the Creative Commons Attribution 4.0 International (CC BY 4.0) license. The dataset is delivered as a single folder named “iraq_prostate_wsi”, which contains six subfolders: three component folders with the WSIs from each scanner, an annotations folder with Gleason score and ISUP grade assignments from three pathologists, a “file_lists” folder documenting cross-scanner image pairing, and a "docs" folder with the dataset documentation. A detailed description of these components is provided in the Data Records section. The complete dataset is deposited in BIA under accession number S-BIAD2323 and can be accessed at: (https://identifiers.org/biostudies:S-BIAD2323).

## Code availability

No custom code was used to generate or process this dataset. Slides were produced with standard laboratory scanner software. No analysis scripts or preprocessing pipelines are required for reuse.

## Author contributions

Study conception and design, software and data handling, operational workflows, and manuscript drafting: P.J.M.A., N.V., S.S.A., H.N.M.F., K.K., A.K.A.-T., N.M. Laboratory and pathology activities and grading: S.S.A. (full slide preparation and grading for all slides), H.N.M.F. (graded 337 slides), A.B. (graded 59 slides). K.S. set up and maintained Cytomine, J.A.M. did scanning and data management, and X.J. did data management and processing of the slides. All authors reviewed the manuscript, and K.K., A.K.A., and N.M. supervised the work.

## Competing interests

K.K. and N.M. are shareholders of Clinsight AB. Other authors declare no competing interests.

## Funding

K.K. received funding from the SciLifeLab & Wallenberg Data-Driven Life Science Program (KAW 2024.0159), the Instrumentarium Science Foundation, and the Karolinska Institutet Research Foundation.

## Acknowledgements

We thank PAR Hospital histopathology lab for their support in preparing slides. High-performance computing was supported by the National Academic Infrastructure for Supercomputing in Sweden (NAISS) and the Swedish National Infrastructure for Computing (SNIC) at C3SE partially funded by the Swedish Research Council through grant agreement no. 2022-06725 and no. 2018-05973, by the supercomputing resource Berzelius provided by the National Supercomputer Centre at Linköping University and the Knut and Alice Wallenberg Foundation.